# Multi-similarity Based Hyperrelation Network for Few-Shot Segmentation

Xiangwen Shi[1,2,*], Zhe Cui[1,*], Shaobing Zhang[1], Miao Cheng[1], Lian He[1], Xianghong Tang[3,*]

**Abstract**

   Few-shot semantic segmentation aims at recognizing the object regions of unseen categories with only a few annotated examples as supervision. The key to few-shot segmentation is to establish a robust semantic relationship between the support and query images and to prevent overfitting. In this paper, we propose an effective Multi-Similarity Hyperrelation Network (MSHNet) to tackle the few-shot semantic segmentation problem. In MSHNet, we propose a new Generative Prototype Similarity (GPS), which, together with cosine similarity, establishes a strong semantic relationship between supported images and query images. In addition, we propose a Symmetric Merging Block (SMB) in MSHNet to efficiently merge multi-layer, multi-shot, multi-similarity features to generate hyperrelation features for semantic segmentation. On two benchmark semantic segmentation datasets Pascal-$5^i$ and COCO-$20^i$, MSHNet achieves new state-of-the-art performances on 1-shot and 5-shot semantic segmentation tasks. Our code is available at https://github.com/Alex-ShiLei/MSHNet .

## 1. Introduction

   The image semantic segmentation aims to classify each pixel of an image into a specific class. It is important fundamental processing in medical analysis and industry defect detection. With the great success of deep learning, image semantic segmentation has made great progress, such as FCN [4], U-Net [7], DeepLab [9], and so on. But training these segmentation models requires a lot of pixel-wise annotated images, which is tedious and costly. Moreover, a trained model can only make predictions within a set of pre-defined classes.

   To perform semantic segmentation for new categories with only a few labeled samples, many few-shot semantic segmentation methods [24,25,26,27] have been proposed. Most of the few-shot segmentation methods are based on two-branch architecture, as shown in Fig 1, with a support branch and a query branch. Support images are labeled with corresponding categories, and query images are segmented according to the labels of the support image. The two-branch structure uses a pre-trained convolutional neural network (CNN) to extract the high-level features of support and query images, and then uses a relational transformation block to establish the semantic relationship between support and query images to perform semantic segmentation on query images. The key to few-shot semantic segmentation is to find the relationship between support images and query images in high-dimensional features.

   There are two ways to establish the semantic relationship between support and query images. The first method [24,25] is that concatenate support features with query feature-map in deep layers to generate the relationship. Another way [26,27] is to seek the cosine similarity between the deep query feature and support feature as a prior probability, and then concatenate it with the query feature map to segment the query image. However, objects are diverse, and it is difficult to describe the relationship between the query and support images robustly using only just one relational function. The HSNet [30] proposed by Min J et al. tells us that using more relations between feature layers is helpful to establish a robust relationship between support and query images.

   Motivated by these, we propose a Multi-similarity Hyperrelation Network (MSHNet) to address few-shot semantic segmentation. First, we use generative prototype similarity and cosine similarity to construct a multi-layer and multi-scale high-dimensional spatial relationship between the query and support images. Compared with feature-based methods [24,38], our prototype-similarity is universal and can generalize models to new categories better. We then use Symmetric Merging Block (SMB) to merge these multiple relationships and establish robust hyperrelation for the support and query images. The locally generated prototype similarity based on global feature is logically complementary to the global cosine similarity based on local feature, and the relationship between the query image and the supported image can be expressed more comprehensively by using the two similarities simultaneously. In summary, MSHNet is based on the similarity relationship, rather than the features of the query image, which reduces the impact of the bias between train and test classes. In addition, the network efficiently integrates the multi-layer and multi-scale relationship features, which can better solve the problem of the diversity of object shapes and sizes. We experiment with the proposed method on Pascal-$5^i$ dataset and COCO-$20^i$ dataset, and we achieve a new the-state-of-art on both datasets. Through experiments, we find that even if we only use one of the two similarities, we can get results close to the existing optimal methods. The contributions of this paper can be summarized as follows.

- We propose a new prototype similarity method, which can effectively describe the similarity between query features and support vectors.
- We use multiple similarities to establish a more robust relationship between support and query images.
- We proved through experiments that the proposed prototype similarity and cosine similarity are complementary.
- The proposed Symmetric Merging Block (SMB) module is not only simple but also efficiently merges the similarity feature of multi-layer, multi-shot, and multi-similarity.
- MSHNet achieves a new state-of-the-art performance on both Pascal-$5^i$ [1] and COCO-$20^i$ [32] datasets.

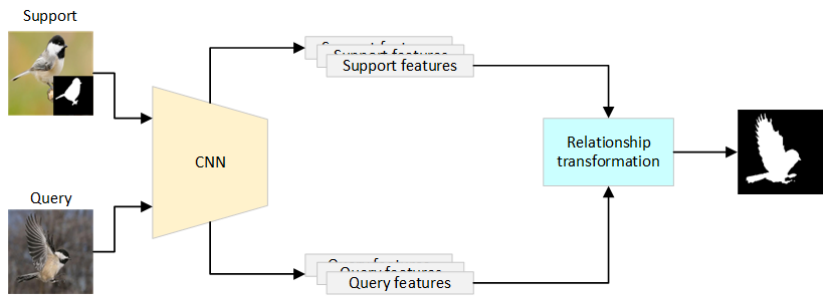

Fig.1 Two branches few-shot semantic segmentation structure.

## 2. Related Work

In this chapter, we briefly review some work related to this paper, such as few-shot learning, image semantic segmentation, few-shot semantic segmentation and so on.

**2.1 Image Semantic Segmentation**

Image semantic segmentation is an essential component in many visual understanding systems, such as autonomous vehicles, and medical image diagnostics. The goal of image semantic segmentation is to classify each pixel of an image into a set of predefined semantic classes. Long et al. [4] proposed fully convolutional networks which trained end-to-end, and pixels-to-pixels on semantic segmentation exceed the previous best results. Encoder-Decoder [5,6,7] is another popular semantic segmentation architecture. The encoder uses convolution and down-sampling to obtain deep features of different scales of the input image, while the decoder merges and upsamples these features to obtain the final semantic segmentation result. The DeepLab family [8,9,10] uses atrous convolution and atrous spatial pyramid pooling (ASPP) to merge the multi-scale information, which can address the decreasing resolution in the encoder and robustly segment objects at multiple scales. Although these methods can achieve image semantic segmentation well, they need a large number of pixel-level annotated images to supervise the network, and the trained model cannot be used for new categories.

**2.2 Few-shot Learning for Image Classification**

Few-shot image classification [11,12,13,14] aims to classify new images with a few examples. Meta-learning [3,19,20,21,22,23] is a common method to solve the few-shot problem. It iteratively trains few-shot tasks on the train set to learn meta-knowledge and adapt the model to the situation with only a few samples. Many meta-learning methods have been proposed based on metric learning and optimization algorithms. Metric-based methods [15,16,17,18,19] aim to learn a general metric function to infer the distances between the query image and samples. The optimization-based [20,21,22,23] method aims to design effective learning strategies for few-shot tasks. For example, Chelsea et al. [20] proposed a good parameter initialization learning method, which can make the model adapt to new tasks quickly. Most of the few-shot learning methods are developed for other image tasks.

## 2.3 Few-shot Semantic Segmentation

Few-shot segmentation [24,25,26,27,30] is the extension of meta-learning methods, which can perform semantic segmentation within a few labeled samples. This research problem was introduced by Shaban et al. [1], who proposed a classical two-branch network. The most challenge for few-shot segmentation is how to design a transformation block and establish a class-agnostic relationship between the support and query images. The relationship is mainly based on generative similarity and cosine similarity. The generative-based methods [24,25,38] directly concatenate the support and query features into the merge block to estimate the prediction. For example, Zhang et al. [24] upsamples the prototype feature vector from the support set and concatenates it with the query feature for dense predictions. However, the training category is different from the test category. Using a large number of parameters to express the relationship between the support and query features will result in overfitting. In the cosine-based methods, [26,27] compute the cosine similarity between the prototypes vector of support feature and pixels of query feature, [28,41] introduce a graph attention mechanism to establish pixel-to-pixel similarity between support and query features. However, these models only use a few feature layers, so it is difficult to establish a robust relationship between the support and query features. Recently, [30,46] used 4D convolution to establish the hyperrelation between multi-layer features, but 4D convolution has high spatial complexity and time complexity. In this paper, we use multi-similarity to build a more robust semantic relationship between support and query images.

## 3. Problem Description：

The few-shot segmentation aims to learn a model to perform segmentation on novel classes with a few pixel-level annotated samples. In the K-shot segmentation task, each class has K pixel-level labeled samples. Suppose we have two non-overlapping data sets $D_{train}$ and $D_{test}$. The train set $D_{train}$ is constructed by classes $C_{train}$ and the test set $D_{test}$ is constructed by $C_{test}$, where $C_{train}$ and $C_{test}$ are non-overlapping ( $C_{train} \cap C_{test} = \varnothing$ ). The model only uses the $D_{train}$ to train the network, and the categories in the $C_{test}$ are not involved in the training process.

Following the previous few-shot methods [2,30], we train and test the model in episodes. In each episode, we randomly sample a support set $S = (X_s, M_s)$ and a same class query set $Q = (X_q, M_q)$, where $X \in R^{3 \times H \times W}$ indicates the RGB image and $M \in \{0,1\}^{H \times W}$ is the corresponding binary segmentation masks. The support set S consists of N classes and each class contains K samples $S = \{(x_c^1, m_c^1)...(x_c^K, m_c^K)\}_{c=1}^N$. In this paper, we train and tested on 1-shot and 5-shot tasks. We are supposed to learn a mapping function $\hat{M} = f_\theta \{X_q | (X_s, M_s)\}$ on $D_{train}$ and evaluate on $D_{test}$, where $\theta$ is the parameters of the model and $\hat{M}$ is the predicted mask. Once the mapping function is learned, the parameters in the model are fixed and we do not optimize the model on the $D_{test}$.

## 4. Our Method Overview

We propose a novel few-shot semantic segmentation architecture, Multi-similarity Hyperrelation Network (MSHNet), which is based on generative prototype similarity and cosine similarity. The overall architecture is illustrated in Fig 2, which is built upon a two-branch architecture. The support branch and query branch share a

CNN block (pretrained on ImageNet [41]) which is used to obtain the high-dimensional semantic features of the input images. We then used these high-level features to calculate the prototype similarity and cosine similarity between the support and query images. Each symmetrical merging block (SMB) takes the prototype similarity and cosine similarity as input to generate hyperrelation features for semantic segmentation. To reduce the risk of overfitting, we train the model in episode [34]. The overview of the whole process is provided in Algorithm 1.

In each episode, we first input the random sampled $X_s \in R^{3 \times H \times W}$ and $X_q \in R^{3 \times H \times W}$ to support and query branch respectively. The pre-trained CNN block extracts multi-scale support deep features $F_s^{b,l} \in R^{d_l \times h_b \times w_b}$ and query deep features $F_q^{b,l} \in R^{d_l \times h_b \times w_b}$, where $d_l$ is the feature dimensionality of layer $l$, and $w_b, h_b$ denote the width and height of feature maps in the block $B$. Then we apply the mask to each support feature map to get object features $F_s^{b,l,o}$ and use average pooling to obtain the class prototype vector $P_s^{b,l,c}$ of the annotated object, which is described in section 4.1. Next, we apply CS() to get pixel-to-pixel cosine similarity between $F_s^{b,l,o}$ and $F_q^{b,l}$, and apply GPS() to get prototype similarity between $P_s^{b,l,c}$ and $F_q^{b,l}$, which is described in section 4.2. In the end, the transformation block merges the multi-layer and multi-scale features, as described in section 4.3. In this paper, we use ResNet-50 [31] and ResNet-101 [31] as the CNN backbone to extract the feature of support images and query images.

We select the standard cross-entropy loss $L(\hat{m}_q, m_q)$ as our loss function. As shown in Fig.2, Each symmetric merging block of the transformation block has an intermediate output and an intermediate loss $L_{inner}^i$ ($i \in \{1,2,3\}$) to help optimize the network. The final prediction of MSHNet generates another loss function $L_{out}$. The total loss L is the sum of $L_{inner}$ and $L_{out}$ as:

$$L = \frac{1}{n} * \sum_{i=1}^{n} L_{inner}^i + L_{out} \qquad (1)$$

where n is the number of symmetric merging blocks. The output of each symmetric merging block is segmentation at a specific scale, and the final output of the network merges the segmentation results at a variety of scales. The model parameters are only updated during training and fixed during testing.

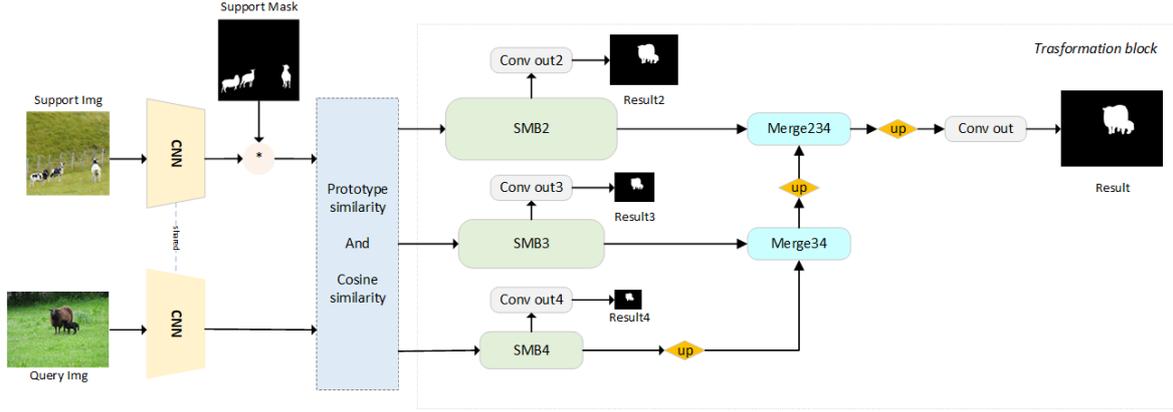

Fig.2 Overview of MSHNet framework. We extract image features by Resnet-50 or Resnet-101 in multi-scale and layers. In each symmetric merging block, we get the layer relationship of the same scale between the support image and query image. Finally, we merge the hyperrelational features of different scales with pyramid structures.

### 4.1 Object Feature Extractor and Prototype

The ResNet [31] is composed of many layers, and each convolutional layer has a specific convolution kernel to generate feature maps. In general, the more pairs of intermediate feature maps $\{(F_s^{b,l}, F_q^{b,l})_{l=0}^{L}\}_{b=1}^{B}$ have been used, the relationship between support and query image is more robust. To balance computational complexity and relational complexity, we only select the features marked in the support image to calculate similarity, as

$$F_s^{b,l,o} = \left\{ F_s^{b,l} \left[ M_s^{b,l}(x,y) = c \right] \right\}_{(x=1, y=1)}^{(x=h, y=w)} \quad (2)$$

where (x, y) is the spatial position of the feature map; [] is a selector function, if true, the index value is selected, otherwise discarded; $F_s^{b,l,o} \in R^{n \times d_l}$. The prototype vector of the specific class object is computed by:

$$P_s^{b,l,c} = \begin{cases} \dfrac{\sum_{i=0}^{|F_s^{b,l,o}|} F_s^{b,l,o,i}}{|F_s^{b,l,o}|} & \text{if } |F_s^{b,l,o}| > 0 \\ 0 & \text{otherwise} \end{cases} \quad (3)$$

where |*| means the number of pixel features in $F_s^{b,l,o}$; $F_s^{b,l,o,i}$ is the i-th value in the support feature set.

### 4.2 Prototype Similarity and Cosine Similarity

Because the class of the test set is different from that of the train set, the feature-based methods [27,38,47] tend to be associated with specific category features. We reduce this bias by class-independent similarity, and use only a few parameters in prototype similarity to further reduce the influence of overfitting. We upsample the prototype vector $P_s^{b,l,c}$ to the query feature map and concatenate it to $F_q^{b,l}$, and then we learn a projection function to get the generative prototype similarity (GPS) between prototype and query features, as

$$GPS(x_p, x_q) = sigmoid(W * [x_p \| x_q]) \quad (4)$$

where $x_p = P_s^{b,l,c}$ is the class prototype vector of the annotated object, $x_q$ is a vector in query feature map, and $W \in R^{1 \times 2d}$ is a full connection layer weight. The output of GPS is a one-dimensional feature map, and the value represents the probability that the position is the target class.

Objects in nature come in a variety of shapes and sizes that are difficult to represent with just a global average prototype. We use the cosine similarity to establish a more detailed relation between query and support in pixel by pixel. The cosine similarity between each point in query feature map and support features is defined as below:

$$CS(x_q, x_s) = \frac{\sum_{i=1}^{|F_s^{b,l,o}|} ReLU(\frac{x_q^T \cdot x_{s,i}}{\|x_q\|\|x_p\|})}{|F_s^{b,l,o}|} \quad (5)$$

where $x_q$ is a vector of a point in $F_q^{b,l}$ and $x_s = F_s^{b,l,o}$, $|F_s^{b,l,o}|$ is the number of elements in $F_s^{b,l,o}$. As done in [30,46], we use the ReLU() function to make the network focus only on how the query point is similar to the specific class object in the supporting image, rather than how they are different. The CS() describes the similarity between each point in query feature map and all feature points in the support set.

After the prototypical similarity and cosine similarity of each layer were obtained, we used a symmetrical merge block to refine the two similarities between the same scale layers.

**4.3 Symmetric Merging Block**

We have obtained prototype similarity and cosine similarity for the multi-layers and multi-shots of the support images and query image. To merge the two similarities more consistently and stably, we propose a two-branch symmetric merging block as shown in Figure 3. The input prototype similarity and cosine similarity are one-dimensional feature map, which were calculated from GPS and CS functions respectively. In Block-Conv, a 1x1 convolution layer was used to integrate similarity features from multiple layers, and then a 3x3 convolution kernel was used to optimize similarity features within the local range. In prototype Shot-Conv, we use multi-scale atrous convolution to obtain the similarity features of different scales in a larger range. Since cosine similarity is stable, we only use a 3x3 convolution layer to merge features in cosine Shot-Conv. In Merge-Conv we use the channel-attention mechanism [29] to highlight the layers of similarity we need to focus on. The symmetric merging block (SMB) merges two similarities and outputs hyperrelational features for semantic segmentation.

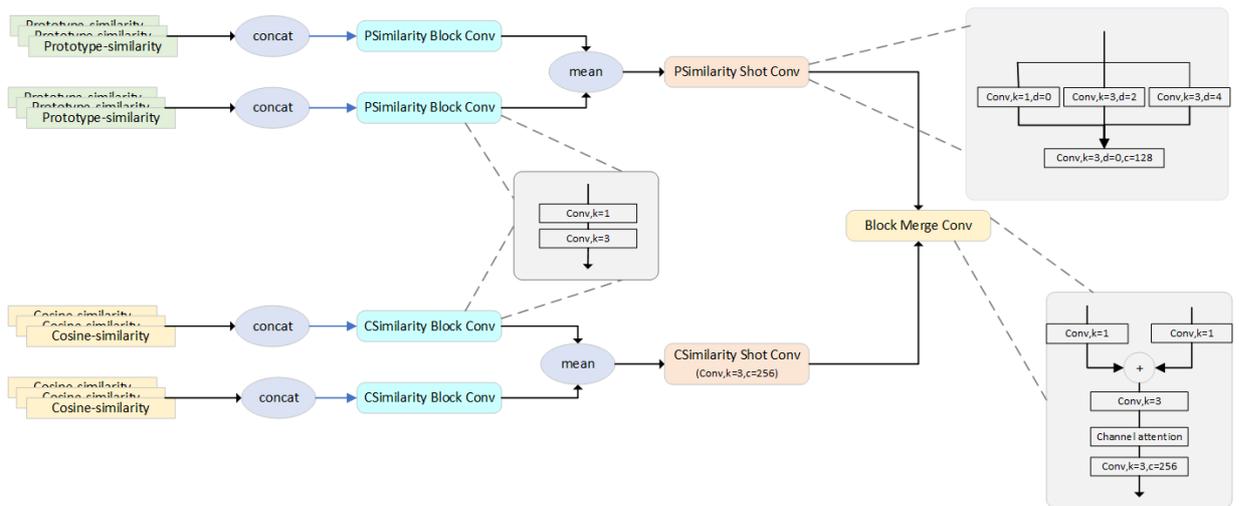

Fig.3 We proposed symmetric merging block. The upper part is used for prototype similarity, and the lower part is used for cosine similarity. Block-Merge-Conv synthesizes these similarity and outputs the hyperrelation features.

**Algorithm 1**: testing an episode for a K-shot semantic segmentation task.

**Input**: Support set S, Support mask $M_s$ and query set Q

Extract query deep features $F_q^{b,l}$ from Q

**for** $b=2,...,B$ **do**

    Extract support deep features $F_{s,k}^{b,l}$ from $s_k$

    **for** $k=1,...,K$ **do**

        **for** $l=1,...,L$ **do**

            Apply $M_{s,k}$ to obtain $F_{s,k}^{b,l,o}$ by Eqn. (2)

            Obtain $P_{s,k}^{b,l,c}$ from $F_{s,k}^{b,l,o}$ by Eqn. (3)

            Compute the generative prototype similarity $GPS_k^{b,l}$ from ($F_q^{b,l}$, $P_{s,k}^{b,l,c}$) by Eqn. (4)

            Compute the cosine similarity $CS_k^{b,l}$ from ($F_q^{b,l}$, $F_{s,k}^{b,l,o}$) by Eqn. (5)

        **end**

        Concatenate [$GPS_k^{b,1}$,…,$GPS_k^{b,L}$] and input it to Psimilarity Block Conv to get $blc\_GPS_k^b$

        Concatenate [$CS_k^{b,1}$,…,$CS_k^{b,L}$] and input it to Csimilarity Block Conv to get $blc\_CS_k^b$

    **end**

    Compute the average of [$blc\_GPS_k^b$,…,$blc\_GPS_K^b$] and input it to Psimilarity Shot Conv to get $shot\_GPS^b$

    Compute the average of [$blc\_CS_k^b$,…, $blc\_CS_K^b$] and input it to Csimilarity Shot Conv to get $shot\_CS^b$

    Input ($shot\_GPS^b$, $shot\_CS^b$) to Block Merge Conv to get hyperrelation $hyper\_sim^b$

**end**

Merge the multi-scale hyperrelational features [$hyper\_sim^b$,.., $hyper\_sim^B$] to get segmentation result.

## 5. Experiment

In this section, we experiment our proposed model on two benchmark datasets and compare the results with current excellent methods.

### 5.1 Pascal-5$^i$

Pascal-5$^i$ using images and annotations in Pascal-VOC 2012[44] and extended annotations from SDS[45]. Following [30], the 20 object categories in Pascal-VOC 2012 dataset is sub-divided into 4 folds i={0,…,3}, each fold having 5 classes $L_{fold}$={4x+i}, where x∈{0,…,4}. Follw [24,28,30,24,35,36], we use cross-validation to evaluate the proposed model, taking the classes in one of folds as test set $C_{test}$ and the classes in the other fold as the train set $C_{train}$. As shown in Table 1, the sample number distribution of different categories in the dataset is very uneven, which will bias the prediction result to the category with more samples. To solve this problem, we randomly select 750 images of each train class to form the train set. Another potential problem is that if objects belonging to the test class appear in the image of the training set, the category will always be predicted as the background during training, and the network will also tend to predict such objects as the background during testing. Therefore, the pictures containing the class of the test set are not used in the training of the network in this paper.

Table.1 Pascle-5$^i$ data set category and corresponding quantity distribution

| Fold0 | Object name | aero-plane | bicycle | bird | boat | bottle |
|---|---|---|---|---|---|---|
| | Number | 528 | 428 | 583 | 371 | 376 |
| Fold1 | Object name | Bus | car | cat | chair | cow |
| | Number | 367 | 848 | 987 | 988 | 235 |
| Fold2 | Object name | dining-tables | dog | horse | motorbike | person |
| | Number | 504 | 1101 | 370 | 440 | 3386 |
| Fold3 | Object name | potted-plant | sheep | sofa | train | tv-monitor |
| | Number | 359 | 276 | 487 | 488 | 476 |

### 5.2 COCO-20$^i$

COCO-20$^i$ is based on the MS COCO2014 dataset [32]. We also use cross-validation to test our model.

Fellowing [30], the 80 classes are divided into 4 folds i={0,1,2,3}, and each contain 20 classes, $L_{fold}$={4x+i}, where x∈{0,1,…,19}. There are ambiguity and overwriting problems when converting JSON-formatted annotation files into grayscale images. For example, if a child wears a dress with a cake pattern, the pattern will be marked as both a person and a cake, while the grayscale image can only choose one of two. In this paper, we directly use JSON-format file to train my network. To solve the problem of unbalanced data, we randomly selected the classes in each training epoch, and then randomly selected images according to the selected classes.

### 5.3 Training and Evaluation

Our MSHNet is implemented on PyTorch[33]. We use ResNet-50 and ResNet-101 pre-trained on ImageNet as our CNN backbones. We adopt the same method as in [30] to extract the features of each layer. We resize input spatial sizes of both support and query image to 473 x473, thus we having $H_2$, $W_2$=60, $H_3$, $W_3$=30 and $H_4$, $W_4$=15. The CNN backbone parameters are fixed and other parameters in MSHNet are initialized by the default setting of Pytorch. We use SGD as our optimizer, the weight decay is 0.0005 and the moment is 0.9. We train the network with an initial learning rate of 0.025 and use exponential decay as the learning rate decay strategy. The models are trained on a single Nvidia Tesla P100 GPU with batch size 10. In each fold test dataset, we random sample 1000 support-query pairs of test images from the selected test classes as test data.

### 5.4 Evaluation Metric

Following previous works [24,25,26], we use the mean intersection-over-union(mIoU) and foreground-background intersection over union (FB-IoU) as our quantitative evaluation. IoU is defined as $\frac{TP}{TP+FP+FN}$, where the TP, FP and FN denote the number of true positive, false positive and false negative pixels of the predicted segmentation mask. The mIoU is the average of all classes IoU: $mIoU = \frac{1}{n}\sum_{i=0}^{n} IoU_i$, where n is the number of classes. FB-IoU is the average of foreground and background IoUs, and ignore the specific object class.

### 5.5 Results and Comparision

In table2, we compare MSHNet with the previous state-of-the-art methods on Pascal-$5^i$ dataset. In the previous methods, HSNet also used multi-layer and multi-scale CNN backbone features. Results in table1 show that the MSHNet outperforms the state-of-the-art methods in both 1-shot and 5-shot tasks. In the 1-shot segmentation tasks, ResNet-50 and ResNet-101 were used as the CNN backbone respectively, and our method achieve 1.3% and 0.5% improvement in mIoU. In the 5-shot segmentation task, using ResNet-50 as the backbone, our model outperforms the CyCTR by 4.3% on mIoU, and using ResNet-101 as the backbone, our method outperforms the CyCTR by 5.7% on mIoU. The result of 5-shot task of HSNet [30] method is obtained by multiple 1-shot tasks. In our proposed method, 5-shot segmentation result is obtained directly through the model. Compared to HSNet, with RESNET-50 as the backbone, our network improved 0.4% mIoU and 1.9% mIoU in 1-Shot and 5-Shot tasks,repectively, and with ResNet-101 as the backbone, our method improved 0.4% mIoU and 1.9% mIoU on 1-shot and 5-Shot tasks.

Table 2. Performance of 1-shot and 5-shot segmentation on Pascal-$5^i$ data set. ("-" means the original paper does not report its performance for this metric.)

| Backbone | Methods | 1-shot | | | | | | 5-shot | | | | | | Learnable params |
|---|---|---|---|---|---|---|---|---|---|---|---|---|---|---|
| | | Fold-0 | Fold-1 | Fold-2 | Fold-3 | mIoU | FB-IoU | Fold-0 | Fold-1 | Fold-2 | Fold-3 | mIoU | FB-IoU | |
| ResNet-50 | PANet [34] | 44.0 | 57.5 | 50.8 | 44.0 | 49.1 | - | 55.3 | 67.2 | 61.3 | 53.2 | 59.3 | | 23.5M |
| | CANet [24] | 52.5 | 65.9 | 51.3 | 51.9 | 55.4 | 66.2 | 55.5 | 67.8 | 51.9 | 53.2 | 57.1 | 69.6 | - |
| | RPMMs [38] | 55.2 | 66.9 | 52.6 | 50.7 | 56.4 | - | 56.3 | 67.3 | 54.5 | 51.0 | 57.3 | - | - |
| | PFENet [35] | 61.7 | 69.5 | 55.4 | 56.3 | 60.8 | 73.3 | 63.1 | 70.7 | 55.8 | 57.9 | 61.9 | 73.9 | 10.8M |
| | HSNet [30] | 64.3 | 70.7 | **60.3** | 60.5 | 64.0 | 76.7 | 70.3 | 73.2 | **67.4** | 67.1 | 69.5 | 80.6 | 2.6M |
| | CyCTR [36] | **67.8** | **72.8** | 58.0 | 58.0 | 64.2 | - | 71.1 | 73.2 | 60.5 | 57.5 | 65.6 | - | - |
| | MSHNet (ours) | 67.7 | 71.8 | 59.9 | **62.4** | **65.5** | **77.8** | **71.3** | **74.1** | 65.7 | **68.5** | **69.9** | **81.0** | 29.6M |
| ResNet-101 | FWB [42] | 51.3 | 64.5 | 56.7 | 52.2 | 56.2 | - | 54.8 | 67.4 | 62.2 | 55.3 | 59.9 | - | 43.0M |
| | ASGNet [43] | 59.8 | 67.4 | 55.6 | 54.4 | 59.3 | 71.7 | 64.6 | 71.3 | 64.2 | 57.3 | 64.4 | 75.2 | 10.4M |
| | PFENet [35] | 60.5 | 69.4 | 54.4 | 55.9 | 60.1 | 72.9 | 62.8 | 70.4 | 54.9 | 57.6 | 61.4 | 77.5 | 10.8M |
| | HSNet [30] | 67.3 | 72.3 | **62.0** | 63.1 | 66.2 | 77.6 | 71.8 | 74.4 | 67.0 | 68.3 | 70.4 | 80.6 | 2.6M |
| | CyCTR [36] | **69.3** | 72.7 | 56.5 | 58.6 | 64.3 | 72.9 | **73.5** | 74.0 | 58.6 | 60.2 | 66.6 | 75.0 | - |
| | MSHNet (ours) | 68.6 | **73.1** | 61.3 | **63.9** | **66.7** | **77.7** | 73.0 | **76.3** | **68.1** | **71.9** | **72.3** | **82.7** | 29.6M |

In Table 3, we compare MSHNet with the state-of-the-art methods on the COCO-$20^i$ dataset. For both 1-shot and 5-shot segmentation tasks our model set a new state-of-the-art. Under 1-shot settings, it outperforms the previous state-of-art method 3.8% and 4.9% mIoU with Resnet-50 and ResNet-101 respectively. Under 5-shot setting, our model outperforms the previous state-of-art method 7.9% and 6.5 % mIoU. In summary, the network model proposed by us is very effective in solving the task of few-shot semantic segmentation. The background of the COCO-$20^i$ data set is complex, and a good segmentation result requires more local and global information. Prototype similarity and Cosine similarity complement each other to provide more information.

Table 3. Performance of 1-shot and 5-shot segmentation on COCO-$20^i$ data set.

| Backbone | Methods | 1-shot | | | | | 5-shot | | | | |
|---|---|---|---|---|---|---|---|---|---|---|---|
| | | Fold-0 | Fold-1 | Fold-2 | Fold-3 | mIoU | Fold-0 | Fold-1 | Fold-2 | Fold-3 | mIoU |
| ResNet-50 | PPNet [37] | 28.1 | 30.8 | 29.5 | 27.7 | 29.0 | 39.0 | 40.8 | 37.1 | 37.3 | 38.5 |
| | RPMM [38] | 29.5 | 36.8 | 29.0 | 27.0 | 30.6 | 33.8 | 42.0 | 33.0 | 33.3 | 35.5 |
| | CyCTR [36] | 38.9 | 43.0 | 39.6 | 39.8 | 40.3 | 41.1 | 48.9 | 45.2 | 47.0 | 45.6 |
| | HSNet [30] | 36.3 | 43.1 | 38.7 | 38.7 | 39.2 | 43.3 | 51.3 | 48.2 | 45.0 | 46.9 |
| | MSHNet (ours) | **39.6** | **48.3** | **45.5** | **42.8** | **44.1** | **49.2** | **62.5** | **55.6** | **52.0** | **54.8** |
| ResNet-101 | FWB [39] | 17.0 | 18.0 | 21.0 | 28.9 | 21.2 | 19.1 | 21.5 | 23.9 | 30.1 | 23.7 |
| | PFENet [35] | 36.8 | 41.8 | 38.7 | 36.7 | 38.5 | 40.4 | 46.8 | 43.2 | 40.5 | 42.7 |
| | HSNet [30] | 37.2 | 44.1 | 42.4 | 41.3 | 41.2 | 45.9 | 53.0 | 51.8 | 47.1 | 49.5 |
| | MSHNet (ours) | **41.3** | **49.9** | **46.4** | **46.6** | **46.1** | **51.7** | **62.6** | **55.9** | **53.8** | **56.0** |

### 5.6 Qualitative Results

Fig.4 shows some qualitative results of our models. We can see that our network can accurately segment query images with only a few labeled sample images. In the 1-shot task, our network can segment object categories and backgrounds well. In the 5-shot task, our network can segment object categories finely, even if it was a wire fence. The main reason for this is that prototype similarity provides a lot of details for semantic segmentation.

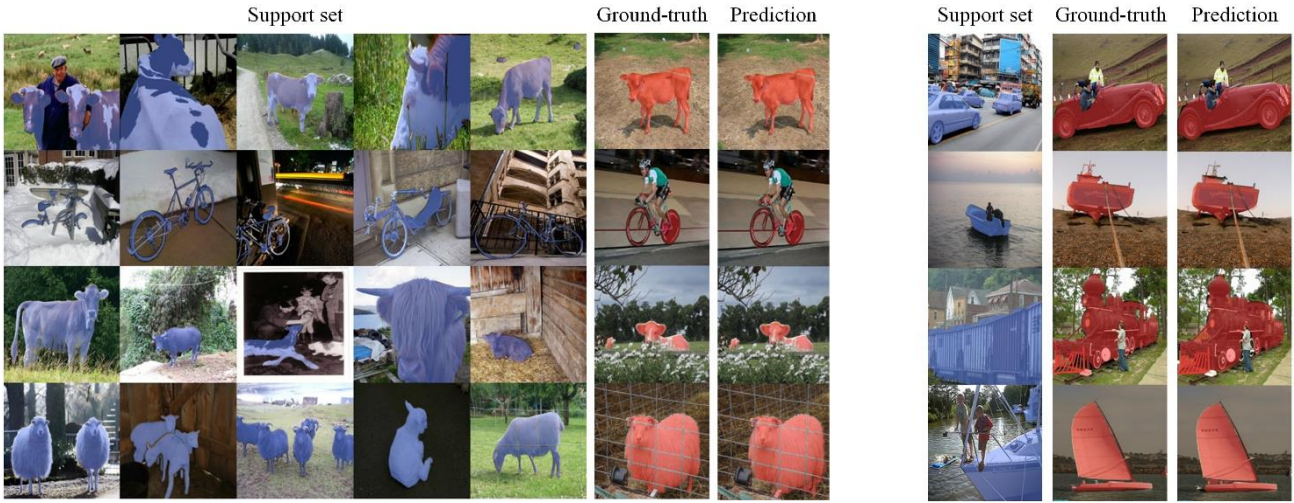

Fig.4. Qualitative results of our network. The left is the result of the 5-shot semantic segmentation and the right is the result of the 1-shot semantic segmentation. Experiments on $Pascal\text{-}5^i$ data set.

## 6. Ablation Study

In this section, we specifically analyze the role of two similarities in the model by analyzing the energy map and using only one similarity for the comparative experiment. Both comparative and analytical experiments were validated on Pascal-$5^i$ data sets.

### 6.1 Energy Map Analysis

The energy map can effectively display the focus of the current convolution layer. We use a pre-trained ResNet as the backbone network to extract high-dimensional features, and calculated the generative prototype similarity and cosine similarity of each feature layer. MSHNet uses all feature maps of the last three blocks of ResNet to calculate the similarity. We use the average of all similarities in one block to represent the energy map, and the strength of the energy value represents the concern of the current layer. From Figure 5, we can see that prototype similarity pays more attention to local content and edge content, and cosine similarity pays more attention to global content. In general, the two similarities are complementary. It can be seen from the energy map that the similarity of deep-level (Block4) features can well represent semantic correlation, but lacks object details. The similarity of low-level features preserves details, but the semantic correlation is not very strong. This indicates that we can use multiple levels of similarity to better segment query images.

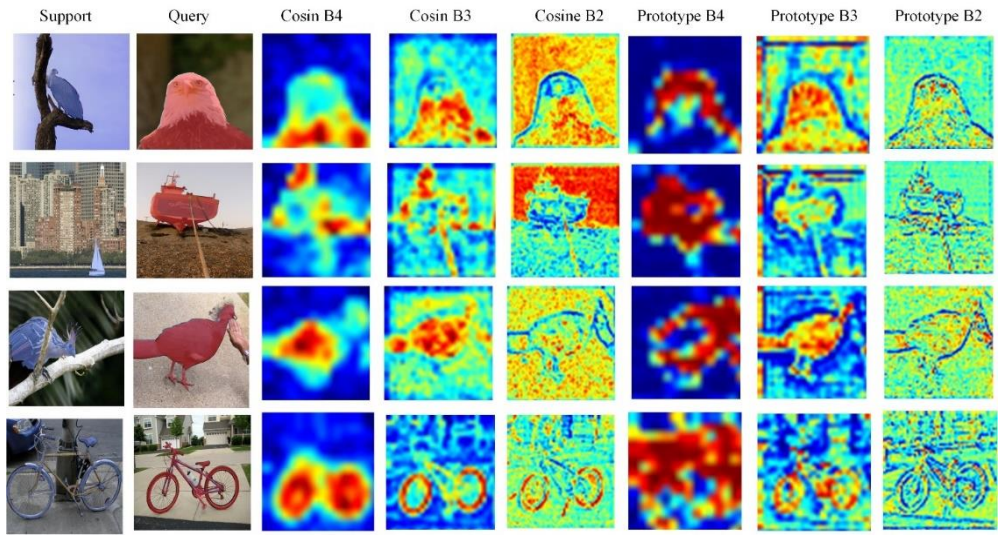

Fig.5 The Block4, Block3, and Block2 energy map of MSHNet.

## 6.2 Test the segmentation effect of single similarity.

In this section, we test the segmentation by using a similarity function alone. We use ResNet-50 and ResNet-101 as the backbone to extract deep features and test the segmentation ability of prototype similarity and cosine similarity respectively. When verifying prototype similarity, we removed the parts only related to cosine similarity from MSHNet, the rest remained unchanged, and so did when verifying prototype similarity. We trained the network with a single similarity. As shown in Table 4, even if only one similarity is used, our network can get a result close to the current best methods. The single-use of either similarity is not as good as the result of the use of both, which shows that the two similarities are complementary.

In Figure 6, we compared the segmentation results of different similarity. As shown in the figure, although not all segmentation results are better than using only one similarity, the combination of two similarities can reduce obvious missegmentation.

Table 4 Results of single similarity on Pascal-5$^i$ data set

| Backbone | Similarity | 1-shot | | | | | 5-shot | | | | | Learnable params |
|---|---|---|---|---|---|---|---|---|---|---|---|---|
| | | Fold-0 | Fold-1 | Fold-2 | Fold-3 | mIoU | Fold-0 | Fold-1 | Fold-2 | Fold-3 | mIoU | |
| ResNet-50 | Cosine | 63.3 | 66.0 | 59.5 | 55.4 | 61.1 | 68.0 | 71.1 | 62.8 | 63.6 | 66.4 | 26.8M |
| | Prototype | 61.6 | 69.2 | 56.5 | 50.5 | 59.5 | 64.3 | 70.6 | 57.4 | 53.4 | 61.4 | 25.9M |
| | Both-Sim | **67.7** | **71.8** | **59.9** | **62.4** | **65.5** | **71.3** | **74.1** | **65.7** | **68.5** | **69.9** | 29.6M |
| ResNet-101 | Cosine | 66.7 | 70.2 | 61.9 | 60.0 | 64.7 | 72.8 | 74.9 | 67.5 | 69.0 | 71.1 | 26.8M |
| | Prototype | 63.4 | 70.3 | 58.3 | 54.6 | 61.7 | 66.2 | 71.3 | 57.7 | 57.2 | 63.1 | 25.9M |
| | Both-Sim | **68.6** | **73.1** | **61.3** | **63.9** | **66.7** | **73.0** | **76.3** | **68.1** | **71.9** | **72.3** | 29.6M |

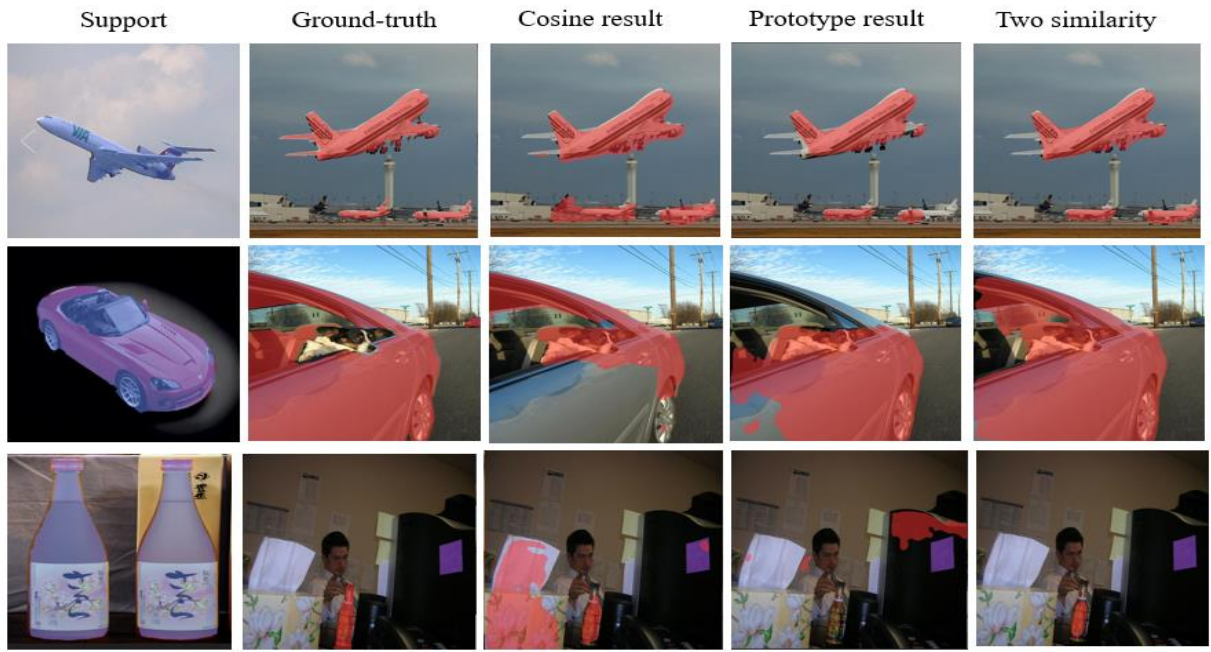

Fig.6 Examples of segmentation with different similarity. We obtained the results with Resnet-101 as the backbone network.

## 7. Conclusion

In this paper, we propose a new few-shot semantic segmentation method called MSHNet, which uses two similarities to find the relation between the support image and query image. The significant improvement in Pascal-5$^i$ and COCO-20$^i$ data sets shows that it is effective to use multiple similarities to address few-shot segmentation problems. According to the energy map, we know that the global feature-based local generative prototype similarity and the local feature-based global cosine similarity are logically complementary, and the relationship between the query image and support image can be more comprehensively expressed by using these two similarities at the same time. The segmentation result of the two similarities is much better than that of the single similarity, which indicates that the symmetric merging block (SMB) proposed by us is very effective.

## 8. discussion

In this paper, we use multi-similarity to solve the few-shot semantic segmentation, and get the best segmentation results. This shows that using multiple similarities can make the relationship between support and query images more robust. But the more similarity a network uses, the more complex it becomes and the more computing resources it requires. Which similarity to choose and how to synthesize these similarities is worth further exploration.

**References**:

[1] Shaban A, Bansal S, Liu Z, et al. One-shot learning for semantic segmentation[J]. arXiv preprint arXiv:1709.03410, 2017.

[2] Vinyals O, Blundell C, Lillicrap T, et al. Matching networks for one shot learning[J]. Advances in neural information processing systems, 2016, 29: 3630-3638.

[3] Wang Q, Liu X, Liu W, et al. Metasearch: Incremental product search via deep meta-learning[J]. IEEE Transactions on Image Processing, 2020, 29: 7549-7564.

[4] Long J, Shelhamer E, Darrell T. Fully convolutional networks for semantic segmentation[C]//Proceedings of the IEEE conference on computer vision and pattern recognition. 2015: 3431-3440.

[5] Noh H, Hong S, Han B. Learning deconvolution network for semantic segmentation[C]//Proceedings of the IEEE international conference on computer vision. 2015: 1520-1528.

[6] Badrinarayanan V, Kendall A, Cipolla R. Segnet: A deep convolutional encoder-decoder architecture for image segmentation[J]. IEEE transactions on pattern analysis and machine intelligence, 2017, 39(12): 2481-2495.

[7] Ronneberger O, Fischer P, Brox T. U-net: Convolutional networks for biomedical image segmentation[C]//International Conference on Medical image computing and computer-assisted intervention. Springer, Cham, 2015: 234-241.

[8] Chen L C, Papandreou G, Kokkinos I, et al. Semantic image segmentation with deep convolutional nets and fully connected crfs[J]. arXiv preprint arXiv:1412.7062, 2014.

[9] Chen L C, Papandreou G, Kokkinos I, et al. Deeplab: Semantic image segmentation with deep convolutional nets, atrous convolution, and fully connected crfs[J]. IEEE transactions on pattern analysis and machine intelligence, 2017, 40(4): 834-848.

[10] Chen L C, Zhu Y, Papandreou G, et al. Encoder-decoder with atrous separable convolution for semantic image segmentation[C]//Proceedings of the European conference on computer vision (ECCV). 2018: 801-818.

[11] Gidaris S, Komodakis N. Dynamic few-shot visual learning without forgetting[C]//Proceedings of the IEEE conference on computer vision and pattern recognition. 2018: 4367-4375.

[12] Dhillon G S, Chaudhari P, Ravichandran A, et al. A baseline for few-shot image classification[J]. arXiv preprint arXiv:1909.02729, 2019.

[13] Yang C, Liu C, Yin X C. Weakly Correlated Knowledge Integration for Few-shot Image Classification[J]. Machine Intelligence Research, 2022, 19(1): 24-37.

[14] Wang Y, Xu C, Liu C, et al. Instance credibility inference for few-shot learning[C]//Proceedings of the IEEE/CVF Conference on Computer Vision and Pattern Recognition. 2020: 12836-12845.

[15] Abdelaziz M, Zhang Z. Multi-scale kronecker-product relation networks for few-shot learning[J]. Multimedia Tools and Applications, 2022: 1-20.

[16] Sung F, Yang Y, Zhang L, et al. Learning to compare: Relation network for few-shot learning[C]//Proceedings of the IEEE conference on computer vision and pattern recognition. 2018: 1199-1208.

[17] Zhang C, Cai Y, Lin G, et al. Deepemd: Differentiable earth mover's distance for few-shot learning[J]. arXiv preprint arXiv:2003.06777, 2020.

[18] Snell J, Swersky K, Zemel R. Prototypical networks for few-shot learning[J]. Advances in neural information processing systems, 2017, 30.

[19]   Mensink T, Verbeek J, Perronnin F, et al. Metric learning for large scale image classification: Generalizing to new classes at near-zero cost[C]//European Conference on Computer Vision. Springer, Berlin, Heidelberg, 2012: 488-501.

[20]   Chelsea Finn, Pieter Abbeel, and Sergey Levine. Model agnostic meta-learning for fast adaptation of deep networks. In International Conference on Machine Learning, pages 1126–1135, 2017.

[21]   Hu S X, Moreno P G, Xiao Y, et al. Empirical bayes transductive meta-learning with synthetic gradients[J]. arXiv preprint arXiv:2004.12696, 2020.

[22]   Rusu A A, Rao D, Sygnowski J, et al. Meta-learning with latent embedding optimization[J]. arXiv preprint arXiv:1807.05960, 2018.

[23]   Li Z, Zhou F, Chen F, et al. Meta-sgd: Learning to learn quickly for few-shot learning[J]. arXiv preprint arXiv:1707.09835, 2017.

[24]   Zhang C, Lin G, Liu F, et al. Canet: Class-agnostic segmentation networks with iterative refinement and attentive few-shot learning[C]//Proceedings of the IEEE/CVF Conference on Computer Vision and Pattern Recognition. 2019: 5217-5226.

[25]   Zhao Y, Price B, Cohen S, et al. Objectness-Aware Few-Shot Semantic Segmentation[J]. arXiv preprint arXiv:2004.02945, 2020.

[26]   Zhang X, Wei Y, Yang Y, et al. Sg-one: Similarity guidance network for one-shot semantic segmentation[J]. IEEE Transactions on Cybernetics, 2020, 50(9): 3855-3865.

[27]   Li G, Jampani V, Sevilla-Lara L, et al. Adaptive prototype learning and allocation for few-shot segmentation[C]//Proceedings of the IEEE/CVF Conference on Computer Vision and Pattern Recognition. 2021: 8334-8343.

[28]   Wang H, Zhang X, Hu Y, et al. Few-shot semantic segmentation with democratic attention networks[C]//European Conference on Computer Vision. Springer, Cham, 2020: 730-746.

[29]   Woo S, Park J, Lee J Y, et al. Cbam: Convolutional block attention module[C]//Proceedings of the European conference on computer vision (ECCV). 2018: 3-19.

[30]   Min J, Kang D, Cho M. Hypercorrelation squeeze for few-shot segmentation[C]//Proceedings of the IEEE/CVF International Conference on Computer Vision. 2021: 6941-6952.

[31]   He K, Zhang X, Ren S, et al. Deep residual learning for image recognition[C]//Proceedings of the IEEE conference on computer vision and pattern recognition. 2016: 770-778.

[32]   Lin T Y, Maire M, Belongie S, et al. Microsoft coco: Common objects in context[C]//European conference on computer vision. Springer, Cham, 2014: 740-755.

[33]   Paszke A, Gross S, Massa F, et al. Pytorch: An imperative style, high-performance deep learning library[J]. Advances in neural information processing systems, 2019, 32.

[34]   Wang K, Liew J H, Zou Y, et al. Panet: Few-shot image semantic segmentation with prototype alignment[C]//Proceedings of the IEEE/CVF International Conference on Computer Vision. 2019: 9197-9206.

[35]   Tian Z, Zhao H, Shu M, et al. Prior guided feature enrichment network for few-shot segmentation[J]. IEEE transactions on pattern analysis and machine intelligence, 2020.

[36]   Zhang G, Kang G, Yang Y, et al. Few-shot segmentation via cycle-consistent transformer[J]. Advances in Neural Information Processing Systems, 2021, 34.

[37]   Liu Y, Zhang X, Zhang S, et al. Part-aware prototype network for few-shot semantic segmentation[C]//European Conference on Computer Vision. Springer, Cham, 2020: 142-158.

[38]   Yang B, Liu C, Li B, et al. Prototype mixture models for few-shot semantic segmentation[C]//European Conference on Computer Vision. Springer, Cham, 2020: 763-778.

[39]   Nguyen K, Todorovic S. Feature weighting and boosting for few-shot segmentation[C]//Proceedings of the IEEE/CVF International Conference on Computer Vision. 2019: 622-631.

[40]   Krizhevsky A, Sutskever I, Hinton G E. Imagenet classification with deep convolutional neural networks[J]. Advances in neural information processing systems, 2012, 25.

[41] Wu Z, Shi X, Lin G, et al. Learning meta-class memory for few-shot semantic segmentation[C]//Proceedings of the IEEE/CVF International Conference on Computer Vision. 2021: 517-526.

[42] Nguyen K, Todorovic S. Feature weighting and boosting for few-shot segmentation[C]//Proceedings of the IEEE/CVF International Conference on Computer Vision. 2019: 622-631.

[43] Li G, Jampani V, Sevilla-Lara L, et al. Adaptive prototype learning and allocation for few-shot segmentation[C]//Proceedings of the IEEE/CVF Conference on Computer Vision and Pattern Recognition. 2021: 8334-8343.

[44] Everingham M, Winn J. The pascal visual object classes challenge 2012 (voc2012) development kit[J]. Pattern Analysis, Statistical Modelling and Computational Learning, Tech. Rep, 2011, 8: 5.

[45] Hariharan B, Arbeláez P, Girshick R, et al. Simultaneous detection and segmentation[C]//European conference on computer vision. Springer, Cham, 2014: 297-312.

[46] Hong S, Cho S, Nam J, et al. Cost Aggregation Is All You Need for Few-Shot Segmentation[J]. arXiv preprint arXiv:2112.11685, 2021.

[47] Li X, Wei T, Chen Y P, et al. Fss-1000: A 1000-class dataset for few-shot segmentation[C]//Proceedings of the IEEE/CVF conference on computer vision and pattern recognition. 2020: 2869-2878.